\documentclass[10pt,conference]{IEEEtran}
\usepackage{amsmath,amssymb,amsfonts}
\usepackage{algorithmic}
\usepackage{graphicx}
\usepackage{textcomp}
\usepackage{xcolor}
\usepackage[hyphens]{url}
\usepackage{fancyhdr}
\usepackage{hyperref}

\usepackage[numbers,sort]{natbib}
\usepackage{booktabs}
\usepackage{multirow}
\usepackage{epigraph}
\usepackage{makecell}
\usepackage{stackengine,scalerel}

\title{An Inquiry into Datacenter TCO\\ for LLM Inference with FP8}
\def\hpcacameraready{} % Uncomment to build camera-ready version

\newcommand\hpcaauthors{Jiwoo Kim$^{1*}$, Joonhyung Lee$^{2*}$, Gunho Park$^{2}$, Byeongwook Kim$^{2}$, Se Jung Kwon$^{2}$, \\ Dongsoo Lee$^{2}$, Youngjoo Lee$^{3}$}
\newcommand\hpcaaffiliation{$^{1}$POSTECH, $^{2}$NAVER Cloud, $^{3}$KAIST}
\newcommand\hpcaemail{jiu.0210@postech.ac.kr, joonhyung.lee@navercorp.com, youngjoo@kaist.ac.kr}

\author{
  \ifdefined\hpcacameraready
    \IEEEauthorblockN{\hpcaauthors{}}
      \IEEEauthorblockA{
        \hpcaaffiliation{} \\
        \hpcaemail{}
      }
  \else
    \IEEEauthorblockN{\normalsize{HPCA \hpcayear{} Submission
      \textbf{\#\hpcasubmissionnumber{}}} \\
      \IEEEauthorblockA{
        Confidential Draft \\
        Do NOT Distribute!!
      }
    }
  \fi 
}

\fancypagestyle{camerareadyfirstpage}{%
  \fancyhead{}
  
  \fancyfoot[C]{}
}
\begin{document}
\maketitle

%Enables the camera ready header and footer
\ifdefined\hpcacameraready 
  \thispagestyle{camerareadyfirstpage}
  \pagestyle{empty}
\else
  \thispagestyle{plain}
  \pagestyle{plain}
\fi

\newcommand{\hpcaheight}{0mm}
\ifdefined\eaopen
\renewcommand{\hpcaheight}{12mm}
\fi

%%%%%%%%%%%%%%%%%%%%%%%%%%%%%%%%%%%%%%%%
%%%%%%%% -- PAPER CONTENT STARTS -- %%%%%%%%%

\begin{abstract}

As large language models (LLMs) continue to scale, the high power consumption of AI accelerators in datacenters presents significant challenges, substantially increasing the total cost of ownership (TCO) for cloud service providers (CSPs) that provide LLM inference.
In this work, we analyze the computational characteristics of LLM inference from a TCO perspective and present a generalizable framework to compare AI accelerators across diverse operational requirements. Using this model, we investigate key workload characteristics influencing TCO for AI accelerators from Intel (Gaudi 2 \& 3) and NVIDIA (H100 \& H200), especially thin GEMM utilization and FP8 quantization.
In particular, as FP8 emerges as the baseline precision for next-generation LLMs, understanding how different architectures implement and benefit from low-precision computation is increasingly critical.
Throughput on thin GEMMs has a greater impact on TCO than theoretical hardware peak throughput because the memory-bound decode phase is dominated by GEMV-like computations. 
We find that Gaudi HPUs achieve superior utilization on thin GEMMs compared to their counterparts, especially in FP8-quantized models. 
Our result underscores the importance of empirical, workload-level analysis in evaluating accelerator performance, rather than relying solely on theoretical hardware specifications.
By studying the interaction between power consumption, quantization strategies, and hardware architecture, we provide insights to support informed deployment decisions and guide future accelerator designs aimed at improving the TCO of LLM inference workloads.

% We find that Gaudi HPUs have superior utilization on thin GEMMs compared to their counterparts, particularly for FP8-quantized models, highlighting the importance of empirical, workload-specific analysis over relying solely on hardware specifications.

\end{abstract}
\section{Introduction}\label{sec:intro}

The rapid advance of large language models (LLMs) has made AI datacenters essential infrastructure for timely and reliable inference workloads. As ever more users submit longer and more complex requests, the number of generated tokens has risen sharply.
The financial and energy burden of deploying LLMs at scale has reached the point where hyperscalers are planning colossal datacenters that would require multiple nuclear power plants to operate. For instance, a single phase of the Stargate project plans to add 4.5\,GW of datacenter capacity\,\citep{stargate_announcement,stargate_oracle}, while Meta has separately announced plans for a 2\,GW datacenter\,\citep{meta_datacenter}. To put these figures into perspective, a typical nuclear power plant generates approximately 1\,GW \citep{nuclear_reactor} and New York City had a peak electricity consumption of 10.8\,GW in 2024 \cite{nyc_power}.

Among the various workloads in the AI pipeline, LLM inference exhibits unique characteristics that require specialized analysis. Unlike highly regular training workloads, LLM inference, particularly the decoding phase, is dominated by memory access and is highly latency sensitive. With the increasing popularity of ``reasoning" AI models for both proprietary models, such as ChatGPT o1 \citep{openai2024o1} and Claude 3.7 \citep{claude37_sonnet}, as well as open-weight models such as DeepSeek R1 \citep{deepseekai2025deepseekr1incentivizingreasoningcapability} and Qwen3 \citep{yang2025qwen3technicalreport}, the number of output tokens per query has significantly increased. As a result, the decoding phase has become the primary performance bottleneck, often dominating prefill in both execution time and system resource usage. This decode-heavy behavior is particularly challenging because it stresses memory bandwidth rather than compute throughput, which many accelerators have traditionally been optimized for.

% 기존의 TCO 연구 limitation 및 우리 연구의 차별성
Despite ongoing advances in AI accelerator design\,\citep{micikevicius2022fp8formatsdeeplearning,lee2025fasterinference} and quantization techniques for LLMs\,\citep{lin2024qserve,ashkboos2024quarot,xiao2024smoothquantaccurateefficientposttraining}, relatively little attention has been paid to how these innovations affect the total cost of ownership (TCO) at datacenter scale. 
Moreover, while previous studies \citep{splitwise, griggs2024melangecostefficientlarge} have analyzed throughput-cost trade-offs, they are often limited to specific deployment scenarios.
In this work, we address this gap by developing a holistic TCO model for LLM inference that accommodates a wide range of cost structures and performance requirements, allowing for flexible evaluation across various cloud service provider (CSP) contexts. Our goal is to provide practical guidance for CSPs by enabling comparisons between hardware options that consider both facility costs and target performance levels. In this work, we present a generalizable TCO estimation framework for selecting AI accelerators in datacenters that emphasizes (1) hardware characteristics related to the numerical precision and (2) workload characteristics specific to the LLM inference.

% TCO 분석에서 중심이 되는 것: hardware characteristic related to FP8, workload characteristic
% FP8 사용에 대한 내용
One of the most effective levers for improving both computational efficiency and memory bandwidth in LLM inference is lowering numerical precision.
% 1) FP8이 실제로 많은 모델에 사용되고 있음
For next-generation LLMs, FP8 is emerging as a de facto baseline data precision.
Recent AI accelerators have begun to incorporate dedicated hardware units optimized for FP8 general matrix-matrix multiplication (GEMM), prompting major model providers to adopt FP8 quantization in production-scale models such as DeepSeek-V3 \citep{deepseekai2024deepseekv3technicalreport} and Qwen3 \citep{yang2025qwen3technicalreport}. 
% 2)  FP8을 써야 하는 이유: GPU generation 거듭에 따라 TFLOPS/Watt 개선 별로 없음 -> 소프트웨어적인 접근 필요
% From the perspective of LLM service deployment, FP8-based management yields a more efficient solution than existing BF16-based management with minimal accuracy drop.
While improvements in dense BrainFloat16 (BF16) \citep{bf16} GEMM throughput and memory bandwidth continue in recent GPU generations, gains in energy efficiency are plateauing, as we show in Table \ref{tab:GPU_generation_comparison}.
Our analysis suggests that focusing solely on high-performance hardware can lead to suboptimal datacenter management, highlighting the necessity of adopting software optimizations, especially low-precision computation.

% FP8-based inference에 관련된 내용
% Based on the FP8 precision-based framework, we analyze the LLM inference workload in detail.
% We primarily investigate the accuracy impacts due to the use of FP8.
% While FP8 is frequently described as a floating-point format, it is more accurately characterized as a family of quantization schemes with varying implementation options.
% Our experiments span key design choices, including E4M3 vs. E5M2 formats, stochastic vs. deterministic rounding strategies.

% However, improving hardware specifications alone without understanding the root causes of bottlenecks is insufficient to increase throughput.
While FP8 provides a promising path toward improved inference efficiency, fully leveraging its potential requires a deeper understanding of workload-specific bottlenecks in LLM inference and corresponding architectural support.
The computational characteristics of LLM inference differ considerably from those of training. Even within the inference process, the prefill and decode phases have distinct workloads. While the prefill phase is similar to the forward pass of training, the decode phase is dominated by thin GEMM operations, which are also referred to as skinny GEMMs, that resemble Generalized Matrix-Vector (GEMV) computations. Thin GEMMs exhibit low computational intensity (CI) and lead to reduced model FLOP utilization (MFU) \citep{palm, megatron_lm}. Unlike most prior works, we perform roofline analysis separately for each phase of LLM inference rather than measuring only the output token throughput across different input/output sequence length configurations. TFLOPS values are calculated for disaggregated workloads instead of TTFT and TPOT \citep{splitwise} in mixed workloads, enabling precise comparisons of computational efficiency relative to the peak throughput at various configurations. 

To summarize our contributions:

\begin{enumerate}
\item We provide a simple framework for comparing the relative cost and throughput for different AI accelerators. Although an oversimplification of the true complexities involved, we believe that our model is general enough for end users to find the balance that suits them best by substituting their actual costs and constraints into the variables we define.
\item Despite the growing popularity of FP8 quantization as a method that offers favorable trade-offs between throughput and memory efficiency, prior work has largely overlooked its empirical evaluation, particularly in the context of hardware-specific behavior during LLM inference. To address this gap, we measure FP8 performance across accelerators from Intel (Gaudi 2 \& 3) and NVIDIA (H100 \& H200).
\item Finally, we provide a focused investigation of LLM decoding throughput that goes beyond a simple inspection of hardware specifications. By examining the performance characteristics and bottlenecks of each device within LLM workloads under FP8-quantized conditions, we offer insights that inform both hardware design and deployment strategies.
\end{enumerate}

\section{Exploring the TCO-Throughput Trade-Off}\label{sec:background}

% \epigraph{People who buy and sell chips think about the cost of chips. People who operate data centers think about the cost of operations. ... Our TCO is so good that even when the competitor’s chips are free, it’s not cheap enough.}{\textit{Jensen Huang, Founder \& CEO of NVIDIA}}

\subsection{Understanding the TCO Model}

\begin{figure}[ht]
    \centering
    
    \includegraphics[width=0.5\textwidth]{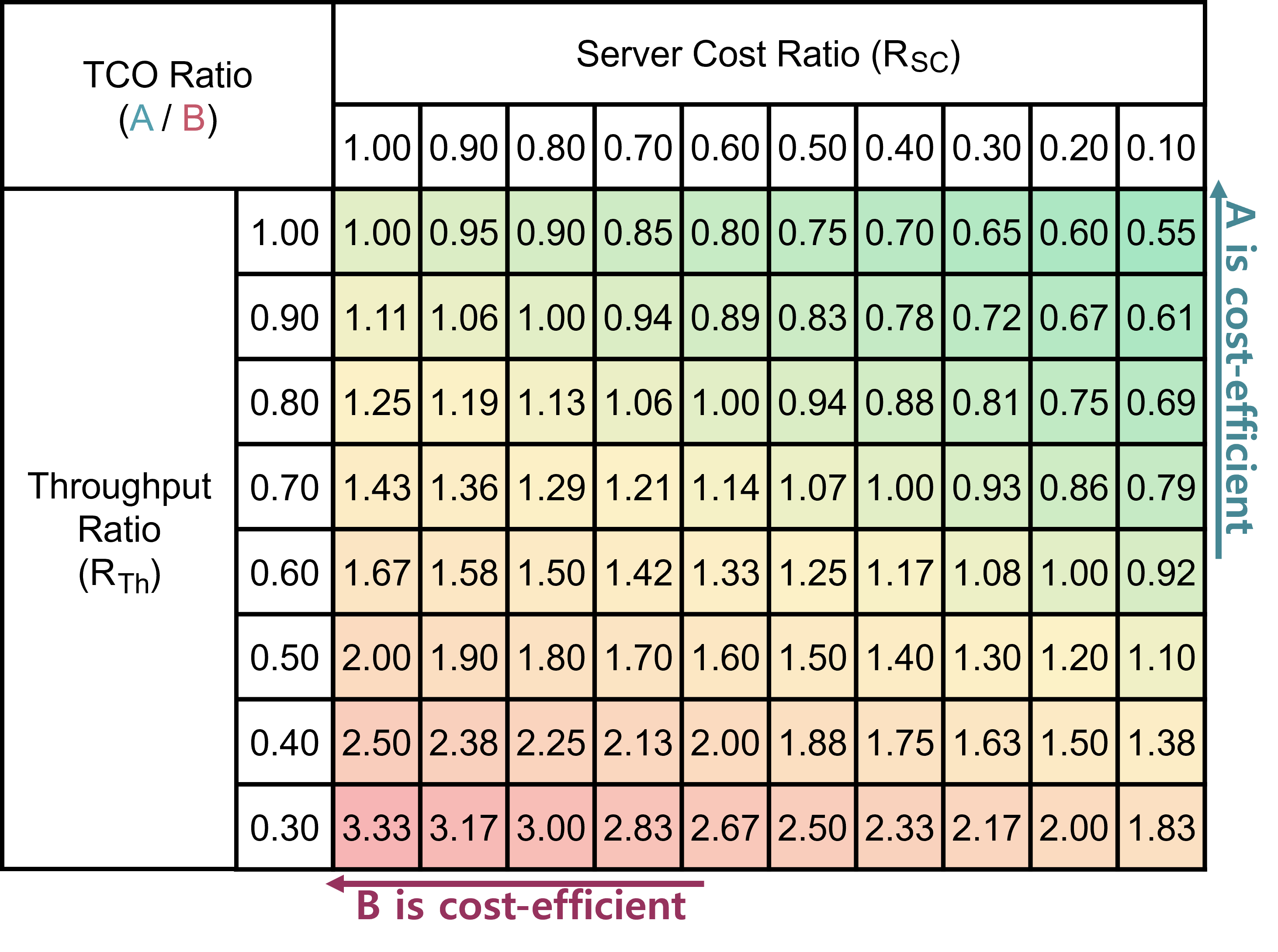}
    \vspace{-1.6em}
    \caption{TCO comparison table for fixed traffic. For any desired task, given the throughput ratio between accelerators A and B $R_{Th}$ and server price ratio $R_{SC}$, the table shows the TCO of accelerator A as a proportion of the TCO of accelerator B. Accelerator A is therefore more cost-efficient than B when the obtained value is less than 1. The lower the value, the greater the cost-effectiveness of A. The values in Equation \ref{eq:tco_iso_traffic} are calculated by assuming ${Cost}_{Server,B}={Cost}_{Infra,B}={Cost}_{Infra,A}$.}
    \label{fig:tco}
    \vskip -0.15in
\end{figure}

% 여기에 TPU v1 논문에서 언급된 TCO 내용을 인용 및 설명하면 좋을 것 같습니다.

When calculating the costs involved in running AI accelerators for datacenter operators, the purchase price of the chips, while significant, constitutes only part of the TCO \citep{tpu_v1}. For example, the purchase cost of the networking, cooling, and power infrastructure necessary to maintain the processors is frequently as large as the purchase cost of the chips and may even exceed it \citep{datacenter_tco_baidu_2017}. As AI inference workloads constitute an ever greater portion of datacenter operations, achieving low cost and high efficiency requires careful planning. The choice of accelerator impacts both the throughput and the power consumption, both of which have a substantial influence on TCO. At large scales, performance per dollar is the most critical metric \citep{tpu_v1}. However, because costs continue to change and often cannot be revealed, we provide a generalizable model where CSPs can enter their own costs and evaluate the results.

Consider a scenario in which a CSP chooses accelerators for a datacenter. Two alternatives exist: accelerator A, which is lighter-weight, delivers lower peak specifications, and has a lower purchase price, and accelerator B, which provides substantially higher peak specifications but comes with a higher per-accelerator purchase price.
To identify which option yields higher profitability, a model to compare TCO values between the two accelerators is needed.
A server contains not only the chip, but also networking, storage, and other associated components. Multiple chips may also be located in a single server. In the following discussion, we denote ${Cost}_{Server}$, abbreviated as ${Cost}_{Sv}$, as the combined purchase costs of the multiple accelerator chips, CPUs, storage, and networking devices required to constitute an operational server.

To clarify the analysis, we use ${Cost}_{Infra}$, abbreviated as ${Cost}_{If}$, to indicate the infrastructure costs associated with operating the datacenter, including cooling and power provision, divided by the number of servers. Operating expenses (OpEx) such as electricity consumption are also included in ${Cost}_{Infra}$. To understand the infrastructure cost, note that while the cost of electricity consumption increases almost linearly with regard to the number of processors, the cost of the associated infrastructure, such as power racks and cooling equipment, is nearly constant. Because datacenter racks can only provide a limited amount of power, the per-chip cost of infrastructure is inversely proportional to the number of servers that fit in a single rack, which in turn depends on the power consumption of the server. Therefore, the benefits of lower power consumption are twofold: the electricity consumption goes down, and the number of processors per rack increases, spreading fixed infrastructure costs among a greater number of servers. For most datacenters, the cost of electricity is outweighed by the cost of the rack and cooling equipment, even when amortized over several years, leading ${Cost}_{Infra}$ to be most heavily affected by the number of chips per rack. We denote $R_{Th}$ as the throughput ratio between a single server of accelerators A and B. We also denote $R_{SC}$ as the cost ratio between accelerators A and B. They are defined as follows:

\begin{align}
    R_{Th} = \frac{{Throughput}_{A}}{{Throughput}_{B}}\\
    R_{SC} = \frac{{Cost}_{Server,A}}{{Cost}_{Server,B}}\\
    R_{IC} = \frac{{Cost}_{Infra,A}}{{Cost}_{Infra,B}}
\end{align}
% The server cost ratio $R_{SC}$, which combines the cost of chips and infrastructure, can be expressed as:

% \begin{equation}
%     R_{SC}=\frac{ChipCost_A+InfraCost_A}{ChipCost_B+InfraCost_B}
% \end{equation}

Regardless of the devices in use, the amount of traffic remains constant. We therefore make an \textit{iso-traffic} assumption and use $N$ to represent the number of B servers required to process the expected traffic. 

Assuming that the cost of infrastructure is the same for accelerators A and B, we derive the following equation:

{\small
\begin{equation}
  \frac{TCO_A}{TCO_B}= \frac{{Cost}_{Sv,B} \times {R_{SC}} \times \frac{N}{R_{Th}} + {Cost}_{If,B} \times {R_{IC}} \times \frac{N}{R_{Th}}} {{Cost}_{Sv,B} \times N + {Cost}_{If,B} \times N}
\label{eq:tco_iso_traffic}
\end{equation}
}

The numerator captures the TCO incurred when scaling up the deployment of lower-performance accelerator A to meet the same throughput target as accelerator B. The denominator reflects the baseline TCO of using accelerator B for a given amount of traffic.
As total traffic remains constant regardless of the underlying device, purchasing accelerator A will require  $R_{Th}/N$ nodes, leading to possibly increased infrastructure costs depending on the throughput ratio $R_{Th}$.
At the same time, its lower per-chip cost, reflected in the server price ratio $R_{SC}$, could result in a lower TCO.

Based on this model, Figure \ref{fig:tco} illustrates the TCO comparison between two accelerators, expressed as an $A/B$ ratio. 
The x-axis represents the server price ratio, and the y-axis represents the per-server throughput ratio of the two accelerators for the desired task. 
Since exact costs are often confidential in real-world data centers \citep{tpu_v1}, allowing users to plug in their costs and flexibly adapt to various operational contexts.
The values in Figure \ref{fig:tco} are an example of the TCO ratio between two hardware options. The lower the value, the less expensive (and hence more cost-effective) lighter-weight accelerator  A becomes, and the more appealing it is to adopt. Conversely, a ratio above 1.0 suggests that higher-weight accelerator B is preferable from a TCO perspective.

\subsection{Visualizing the TCO Model}

% Show how FP8 moves the TCO point in Figure 1. Either use a new figure or explanation. It must show how FP8 moves the TCO point in Figure 1.

% 자연스러운 연결을 위해서 Figure 1에서 FP8이 어떤 영향을 미치는지 설명이 필요하다.

\begin{figure}[ht]
    \centering
     \hspace{-1cm} % 왼쪽으로 1cm 이동
    
    \includegraphics[width=0.5\textwidth]{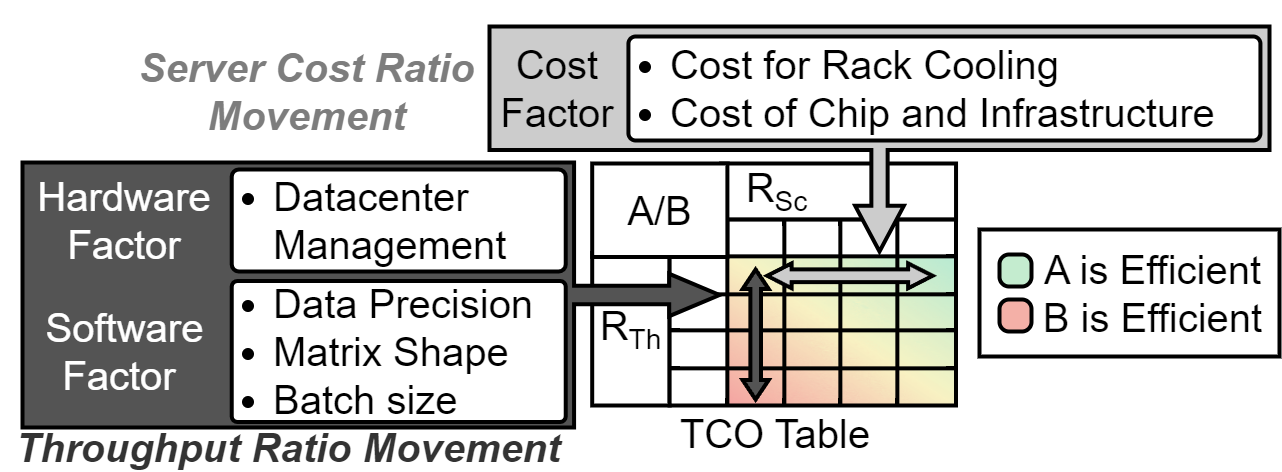}
    \vspace{-1.6em}
    \caption{Visualization of TCO Variation by Server Cost and Throughput Ratios. 
    %The x-axis reflects changes in server cost ratio driven by cost factors, while the y-axis captures throughput ratio variations influenced by hardware and software factors.
    }
    \label{fig:TCO_variation}
    % \vskip -0.1in
\end{figure}

 The TCO table provides interpretability for real-world datacenter environments that serve diverse applications. Figure~\ref{fig:TCO_variation} illustrates how the TCO ratio changes under varying conditions, with a focus on two key parameters: the throughput ratio ($R_{Th}$) and the server cost ratio ($R_{SC}$). 
 %In our model, system A represents Gaudi 2, and system B represents the H100.
Each coordinate in the table represents an evaluation point, where the color indicates the relative cost-efficiency between two accelerators. 

Movement along the x-axis reflects the server cost ratio, which is influenced by device procurement costs, infrastructure constraints, and other datacenter-specific factors. Purchasing accelerator A at a discount would move the TCO ratio along the x-axis to the right. However, even accelerator A was given for free, ${{Cost}_{Server,A}}$ and ${{Cost}_{Infra,A}}$ would both be greater than 0, requiring careful inspection of $R_{Th}$.

% For instance, the cost of mechanical supporting equipment may vary depending on regional climate since the cooling efficiency is affected by the temperature and the moisture of the air \cite{alissa2025using}.
% %while the cost of electrical supporting equipment, such as generators and UPS systems, depends on the stability of the local power grid. 
% These factors are typically fixed for a given datacenter, making the x-axis position relatively static across workloads.

The y-axis position, which reflects the throughput ratio $R_{Th}$, is highly dependent on two operational factors: (1) hardware characteristics related to the data precision and (2) workload characteristics.

As different models and stages of inference have different throughput values on different devices, there is no single throughput value, but many throughput values for each task. For example, increasing the batch size can improve the throughput of a device but comes at the cost of higher latency. The throughput of the accelerator could be defined as the throughput at the largest batch size that meets a specific latency requirement. Likewise, we could also use quantization to shift $R_{Th}$, possibly with the condition that only quantized models that pass specific quality controls may be considered. Accelerators can have different $R_{Th}$ values for different inference phases, e.g., prefill and decode. The next sections discuss these topics in greater detail.

\section{FP8 for High-Throughput LLM Inference}

\begin{table}[ht]
\caption{Comparison of technical specifications across NVIDIA GPUs}
\small
\centering
\begin{tabular}{l|rrcr}
\toprule[1.2pt]
& \textbf{TFLOPS} & \textbf{TDP} & \textbf{TFLOPS/Watt} & \textbf{Increase} \\
\midrule
\textbf{V100 (SXM)} & 125 & 300 & 0.42 & - \\
\textbf{A100 (SXM)} & 312 & 400 & 0.78 & 87\% \\
\textbf{H200 (HGX)} & 989 & 700 & 1.41 & 81\% \\
\textbf{B300 (HGX)} & 2,250 & 1,200 & 1.88 & 33\% \\
\bottomrule[1.2pt]
\end{tabular}
\vskip -0.1in
\label{tab:GPU_generation_comparison}
\end{table}

\subsection{Motivation for FP8 Quantization}

With the increasing adoption of FP8 GEMM in recent accelerator designs, the format is expected to become a next-generation standard for LLM inference.
While most recent LLMs continue to operate in BF16 precision, lowering precision offers a significant opportunity to boost throughput and reduce the TCO, as long as the accuracy degradation remains within acceptable limits.
Despite these benefits, real-world deployment of low-precision inference remains challenging in practice. This is particularly true for decoder-only models, even when dynamic quantization is applied to activations.
However, a new wave of foundation models such as DeepSeek-V3 and LLaMA 4 has demonstrated the feasibility of FP8 at scale. These models are trained using dynamic fine-grained FP8 quantization, showing that FP8 is a practical alternative to BF16 for achieving high-throughput inference while maintaining the quality of outputs.

Employing FP8 can often yield greater gains for LLM inference than simply adopting newer accelerators with improved memory bandwidth or peak compute performance.
Table~\ref{tab:GPU_generation_comparison} compares key architectural specifications across four generations of NVIDIA GPUs: V100 \citep{volta_whitepaper}, A100 \citep{ampere_whitepaper}, H200 \citep{h100_whitepaper, h200_datasheet}, and the upcoming B300 \citep{blackwell_whitepaper}.
The comparison includes peak dense BF16 (except for the V100, which uses FP16) GEMM throughput in TFLOPS (TFLOPs per second), thermal design power (TDP), and energy efficiency measured in TFLOPS per Watt.
Even relying solely on official specifications, the improvement in dense BF16 TFLOPS per Watt shows diminishing returns. The gains from V100 to A100 are significant, but the improvement from H200 to B300 is noticeably smaller.
This suggests that simply scaling hardware resources is no longer sufficient to keep pace with the growing demand for LLM inference.
The bottleneck increasingly lies in thermal and power constraints, which are exacerbated as more compute and I/O components are packed into limited physical space.
As a result, software and algorithmic optimizations, particularly those involving low-precision formats such as FP8, are essential for the next generation of LLM infrastructure.

However, the effectiveness of low-precision computation depends heavily on the underlying hardware architecture.
Although FP8 may offer up to twice the theoretical throughput compared to BF16, achieving this level of performance in practice is often difficult due to reduced hardware utilization.
As the numerical precision decreases, it becomes harder for the system to keep all units busy, and performance bottlenecks in memory bandwidth, special function units (SFUs, also referred to as multi-function units, or MUFUs), or scheduling pipelines can prevent ideal throughput.
Therefore, the actual benefits of FP8 are closely tied to the ability of the architecture to support low-precision computation efficiently.
To guide future LLM deployment strategies, it is essential to perform TCO analysis that explicitly incorporates FP8-related considerations.
Only by aligning hardware and software design around low-precision computation can datacenter architects fully realize the potential efficiency gains of FP8.

\begin{table}
\centering
\caption{Comparison between different FP8 options}
% \vskip 0.15in
\small
\begin{tabular}{@{}lccc@{}}
\toprule
\multicolumn{1}{l}{\textbf{Model}} & \textbf{Data Type} &  \textbf{Rounding} & \textbf{MMLU} \\ \midrule
\multirow{3}{*}{Llama v3.2}  & BF16 & -   & 46.3\% \\
\multirow{3}{*}{1B Instruct}              & E4M3 & SR  & 45.7\% \\
                                        & E4M3 & RTN & 45.5\% \\
                                        & E5M2 & RTN & 44.5\% \\ \midrule
\multirow{3}{*}{Llama v3.2}  & BF16 & -   & 61.8\% \\
\multirow{3}{*}{3B Instruct}          & E4M3 & SR  & 61.7\% \\
                                        & E4M3 & RTN & 61.6\% \\
                                        & E5M2 & RTN & 60.7\% \\ \midrule
\multirow{3}{*}{Llama v3.1}  & BF16 & -   & 68.8\% \\
\multirow{3}{*}{8B Instruct}           & E4M3 & SR  & 68.3\% \\
                                        & E4M3 & RTN & 68.3\% \\
                                        & E5M2 & RTN & 67.5\% \\ \midrule
\multirow{3}{*}{Llama v3.3} & BF16 & -   & 82.0\% \\
\multirow{3}{*}{70B Instruct}              & E4M3 & SR  & 82.0\% \\
                                        & E4M3 & RTN & 82.0\% \\
                                        & E5M2 & RTN & 82.2\% \\ \bottomrule
\end{tabular}
\vskip -0.1in
\label{tab:e5m2_sr_compare}
\end{table}

%However, the improvement in memory and computation performance of hardware accelerators is slowing down.

\subsection{Differences in Hardware Capabilities}

% While only recent accelerators support dedicated compute units for FP8, the characteristics of FP8 should be analyzed to understand how they affect model accuracy.
Although FP8 is widely regarded as a numerical format, it is more accurately described as a form of quantization rather than a fully standardized datatype such as BF16 \citep{bf16}.
The behavior of FP8 depends strongly on the choice of scaling factor, which can vary in granularity from per-tensor to per-channel or even group-based approaches as in the MXFP8 \citep{mxfp_ocp_standard} format.
To explore these differences, we compare FP8 GEMM implementations on NVIDIA GPUs and Intel Gaudi HPUs, identifying key distinctions that affect both throughput and model accuracy despite all these devices conforming to the FP8 specification \citep{micikevicius2022fp8formatsdeeplearning}.

\textbf{(Binary formats)} Both NVIDIA GPUs and Intel Gaudi HPUs support the E4M3 and E5M2 formats \citep{micikevicius2022fp8formatsdeeplearning}. E4M3 assigns one sign bit, four exponent bits, and three mantissa bits. E5M2 also assigns one sign bit but uses five exponent bits and only two mantissa bits. Previous works proposed using E4M3 for the forward pass, which is more sensitive to precision, and E5M2 for the backward pass, which is more sensitive to outliers. However, DeepSeek-V3 used E4M3 in both the forward and backward passes, showing that a smaller group size for scaling factors can compensate for the smaller value range of E4M3. Alternatives such as E3M4 have also been proposed \citep{MLSYS2024_dea9b4b6}, but have yet to be implemented in commercial hardware.

\textbf{(Accumulation precision)} Hopper GPUs use a 14-bit accumulator for FP8 GEMMs \citep{deepseekai2024deepseekv3technicalreport}, requiring casting to CUDA cores for higher precision. Software optimizations, such as applying FP32 accumulation to only one in four warp-group matrix multiply-accumulate (WGMMA) instructions, reduce error but increase kernel complexity and remain less precise than full FP32 accumulation. In contrast, Gaudi HPUs always accumulate in FP32 \citep{lee2025fasterinference}, ensuring higher numerical precision.

\textbf{(E4M3 range)} The Gaudi 2 follows the IEEE specification for special values, unlike NVIDIA GPUs and the Gaudi 3, which use a single special value representation \citep{noune20228bitnumericalformatsdeep}. This results in seven fewer magnitude representations and a maximum value of 240 for E4M3 in the Gaudi 2, compared to 448 on NVIDIA GPUs. This has been addressed in the Gaudi 3, which uses the same E4M3 format as NVIDIA GPUs. Note that E5M2 follows the IEEE 754 specification for all devices.

\textbf{(Power-of-2 scaling)} Gaudi HPUs allow the modification of floating-point exponent biases to accelerate scaling factor application. The Gaudi 2 supports fixed hardware-accelerated scaling factors of  $2^{-8}, 2^{-4}, 2^{0}, 2^{4}$ for E4M3, while the Gaudi 3 extends this to arbitrary powers of 2 between $2^{-32}$ and $2^{31}$. However, this feature is limited to per-tensor scaling factors.

\textbf{(Stochastic rounding)} During FP8 quantization, stochastic rounding can be applied when converting 16/32-bit floating-point values to FP8, similar to the technique proposed for FP32-to-BF16 conversion in \citep{hlat}. This method is distinct from stochastic rounding applied at the inner-product \citep{doi:10.1137/22M1510819}.
However, we find that stochastic rounding does not meaningfully benefit output quality, as shown in Table \ref{tab:e5m2_sr_compare}.

\textbf{(Sparsity)} NVIDIA GPUs support semi-structured sparsity acceleration, potentially doubling tensor core peak throughput. However, despite attempts to leverage semi-structured sparsity in LLM inference \citep{sparsegpt, wanda_sun2024a}, dense GEMMs remain dominant due to accuracy loss and limited speedups. Gaudi HPUs do not support semi-structured sparsity acceleration.

\textbf{(Quantization granularity)} Both Hopper GPUs and Gaudi 2\&3 support standard FP8 \citep{micikevicius2022fp8formatsdeeplearning}, which has limited support for fine-grained quantization. However, starting with Blackwell \citep{blackwell_whitepaper}, NVIDIA GPUs have adopted the MXFP8 \citep{mxfp_ocp_standard} format, which uses an 8-bit power-of-2 shared scaling factor and a group size of 32.

\subsection{Dynamic vs. Static: A Key FP8 Quantization Choice}
% FP8 scaling factor 설정은 static과 dynamic 방법이 있다
A key consideration in utilizing FP8 is the activation quantization strategy. Weights are always quantized ahead of time, but activations must be quantized online before each GEMM operation. Commercially available AI accelerators require that both input matrices for GEMM operations be in FP8. However, operations such as SwiGLU or residual addition require the inputs to be in 16-bit, and RoPE \citep{rope} requires 32-bit \citep{wang2025when}. Therefore, when multiplying activations with weights in an LLM, the activations are usually in BF16 and must be quantized online to FP8. During the online quantization of the activations, the quantization scaling factors can be obtained either statically or dynamically: calculated beforehand based on calibration data or dynamically during inference time \cite{zhao2024atom, xiao2024smoothquantaccurateefficientposttraining}.

Static scaling factors have the advantage that they do not require an additional pass over the high-precision activation to obtain quantization statistics, e.g., the maximum absolute value of a tensor. However, it has the disadvantage that input activations may have a different distribution from the calibration set, causing performance degradation. LLMs often have outlier values in their activations \citep{sun2024massive}, making it challenging to find static scaling factors that perform well for all inputs. Also, per-token fine-grained activation quantization is only possible for dynamic scaling factor quantization. Because inference batches consist of unrelated sequences, information to distinguish between different tokens in a batch is only available at runtime. Static scaling factors cannot apply different scaling factors to different tokens. In practice, this limits static scaling to per-tensor quantization of the activations. Although various techniques, such as smoothing~\citep{xiao2024smoothquantaccurateefficientposttraining} and rotation-based methods~\citep{liu2024spinquant, ashkboos2024quarot}, can help mitigate output quality degradation, they rely on calibration data and may yield suboptimal scaling factors for unseen inputs. Dynamic scaling enhances output quality by assigning separate scaling factors to each token.  As a result, dynamic quantization remains the preferred approach, and native FP8 foundation models such as DeepSeek-V3 and Llama 4 use dynamic activation scaling factors for both training and inference. This restriction does not apply to the weights, which can apply fine-grained static scaling factors during offline quantization. For example, DeepSeek-V3 and Qwen3 use a group size of $128\times 128$ for their FP8 weights.

\subsection{Comparison Between FP8 Options}

We analyze the impact of applying the E4M3 and E5M2 formats, as well as stochastic rounding (SR), on instruction-tuned LLMs in Table~\ref{tab:e5m2_sr_compare}.
All evaluations were conducted on an Intel Gaudi 2 HPU using instruction-tuned LLaMA models with dynamic FP8 scaling factors. Performance was measured on the MMLU benchmark with 5-shot chain-of-thought (CoT). Our results show that E4M3 consistently outperforms E5M2, which is in line with previous work \citep{MLSYS2024_dea9b4b6}.

Stochastic rounding might be expected to preserve more numerical information during quantization by introducing randomness that avoids systematic rounding bias, which could help maintain model accuracy.
However, our experiments in Table \ref{tab:e5m2_sr_compare} indicate that applying stochastic rounding during FP8 quantization does not lead to meaningful accuracy improvements.
In some cases, it even slightly reduces accuracy, consistent with observations from prior work \citep{hlat}.
These findings suggest that the practical value of stochastic rounding for FP8 inference for instruction-tuned LLMs remains limited.

\section{Phase-Aware Analysis of FP8 LLM Inference}\label{sec:g2_vs_h100}

\begin{figure*}[ht]
    \centering
    \includegraphics[width=\textwidth]{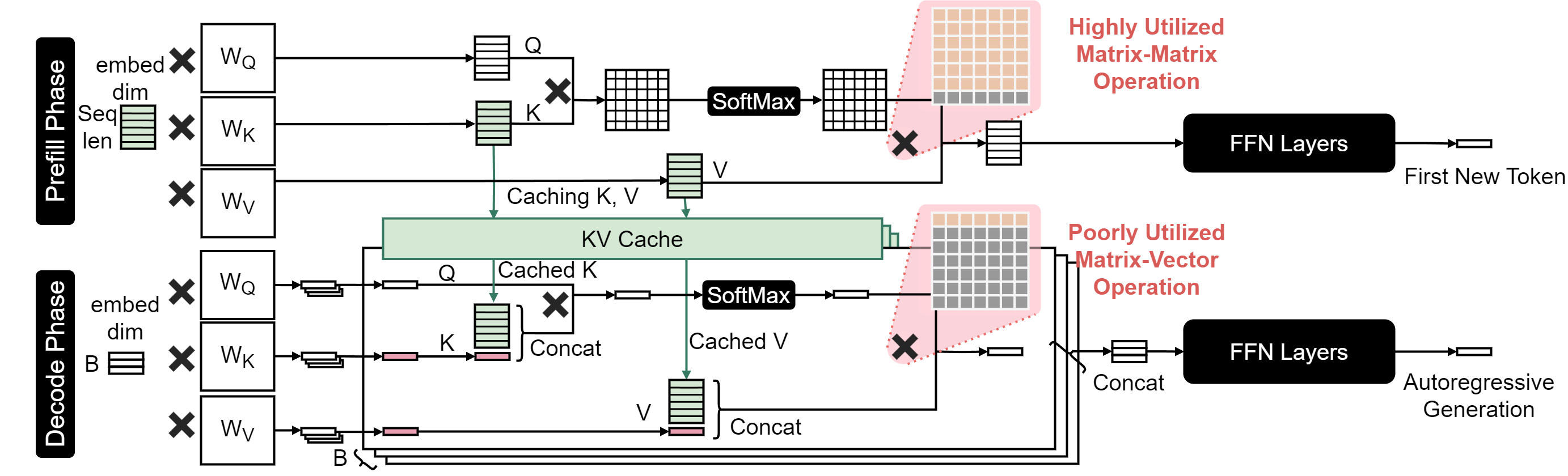}
    \vspace{-1em}
    \caption{Process and utilization characterization of the two phases of generative LLM inference. The prefill phase processes multiple tokens from the same sequence simultaneously, allowing efficient compute-bound computation that produces a single KV cache. The decode stage batches many sequences to improve the computational intensity at the linear layers, but is limited by memory because each sequence has a separate KV cache.}
    \vskip -0.1in
    \label{fig:decode_diagram}
\end{figure*}

% In this section, we explore the characteristics and implications of Llama 4 style FP8 quantization, analyzing its computational behavior and potential for performance optimization. FP8-quantized versions of models trained using BF16 are obtained using post-training quantization (PTQ).

\subsection{LLM Inference Phases}

Generative LLM inference comprises two distinct phases: a compute-bound prefill phase and a memory-bound decode phase \citep{splitwise, distserve_ZhongLCHZL0024}.
Figure~\ref{fig:decode_diagram} illustrates the key differences between these phases. During the prefill phase, the LLM generates the first output token based on the input prompt. The key-value (KV) pairs generated during prefill are stored in the KV cache for reuse in subsequent token generation. The prefill phase processes the entire input in parallel, utilizing large compute-bound GEMMs. This high degree of parallelism enables efficient input processing.

After the first token is generated, the process transitions into the decode phase, where the model generates tokens autoregressively one at a time, conditioning each output on all previously generated tokens.
Unlike the prefill phase, the decode phase is constrained by memory capacity and bandwidth because it involves GEMV and thin GEMM operations, inherently limiting hardware utilization. Inference frameworks such as vLLM \citep{kwon2023efficient} and TensorRT-LLM \citep{TensorRT_LLM} mitigate this issue by batching multiple decoding requests together. This increases the computational intensity (CI) of the linear layers, improving throughput. However, batching is fundamentally limited by memory capacity because each sequence in the batch requires its own KV cache. Moreover, the attention operation during decoding does not benefit from increased batch size. Each sequence still requires a separate GEMV for attention, which prevents CI improvements.
This bottleneck becomes more noticeable at longer sequence lengths.
While the number of FLOPs required for linear layers remains constant, the FLOPs for attention increase linearly with sequence length.
Grouped Query Attention (GQA) \citep{ainslie2023gqa} reduces the overhead by converting the attention computation into a thin GEMM.
Although this increases efficiency, the operation remains memory-bound.
Multi-head Latent Attention (MLA) \citep{deepseekai2024deepseekv2strongeconomicalefficient} further improves the computational intensity of the decode phase, yet the fundamental limitation persists.

\subsection{Calculating GEMM FLOPs}

For a GEMM between matrices of size $(M \times K) \times (K \times N)$, the total number of floating-point operations (FLOPs) performed is $2MKN$. This is derived from the $M \times N$ dot products of length $K$, where each element undergoes one multiplication and one addition.
$M$ corresponds to typical batch sizes during inference and $K,N$ represent hidden dimension sizes. 
Following the convention where FLOPS denotes FLOPs per second, we compute throughput in FLOPS using the theoretical FLOPs and the measured latency. The model FLOPs utilization (MFU) \citep{palm} is derived by dividing the observed throughput by the hardware specification throughput.

\subsection{Calculating Inference FLOPs}

While previous works \citep{splitwise} have evaluated inference performance using time to first token (TTFT) and time per output token (TPOT), these metrics do not facilitate comparisons across different inference stages, model sizes, or sequence lengths. To ensure a consistent comparison, we directly compute model FLOPs.
We follow the methodology described in~\citep{pytorch2, megatron_lm}, computing only the FLOPs associated with matrix multiplications and excluding those related to autoregressive attention masking, which can be skipped \citep{dao2024flashattention}. This yields a more faithful representation of the actual decoding FLOPs.

Using this method, the FLOPs required for a forward pass of a Llama model with $l$ transformer blocks, hidden size $h$, intermediate size $ah$, head size $d$,  head count $H=h/d$, GQA group size $g$, vocabulary size $v$, and input sequence of length $s$ can be calculated as follows:
\begin{equation}
  f_{llama}(s)=2sh^2l(3a+2+\frac{2}{g})+2s^2hl+2vsh.
\label{eq:model_flops}
\end{equation}
By denoting $A=(3a+2+\frac{2}{g})$ as a model-specific constant, Equation~\ref{eq:model_flops} simplifies to:
\begin{equation}
  f_{llama}(s)=2s(Ah^2l+vh)+2s^2hl.
\label{eq:model_flops_simple}
\end{equation}
When the model generates $t$ tokens in a single decoding iteration where $t \ll s$, we approximate the additional computation by evaluating Equation~\ref{eq:model_flops} at $s' = s + t$, yielding:
\begin{equation}
\begin{aligned}
  f_{llama}(s+t)-f_{llama}(s) \\
  \approx 2t(Ah^2l+vh)+4sthl.
\label{eq:model_flops_delta}
\end{aligned}
\end{equation}
From Equation~\ref{eq:model_flops_delta}, we observe that the computational cost of the linear layers, including the LM head, remains independent of the previous sequence length $s$, while the attention-related FLOPs scale proportionally with $s$.
In the autoregressive setting where $t = 1$, and for a batch of $b$ sequences with respective sequence lengths $s_1, \dots, s_b$, the FLOPs required for one decoding step can be approximated as:
\begin{equation}
\begin{aligned}
    2b(Ah^2l+vh)+4hl\sum_{i=1}^{b}s_i.
\label{eq:model_flops_delta_batch}
\end{aligned}
\end{equation}
A key observation is that only the $2bAh^2l$ term, corresponding to the linear layers (excluding the LM head), is computed in FP8. In contrast, the $2bvh$ term for the LM head and the $4hl \sum_{i=1}^{b} s_i$ term for attention are computed in BF16. Additionally, online dequantization of the KV cache introduces extra overhead not accounted for in Equation~\ref{eq:model_flops_delta_batch} as these overheads do not contribute to the model FLOPs as defined in the MFU metric~\citep{palm}.

\subsection{Challenges in Achieving Peak Throughput}

A fundamental limitation in LLM inference is that not every FLOP can be executed at peak efficiency, primarily due to hardware utilization constraints.
For instance, the Gaudi 2 provides a peak HBM bandwidth of 2.4 TB/s and a peak FP8 GEMM throughput of 865 TFLOPS.
This implies that a computational intensity (CI) of at least 360 FLOPs/byte is required to saturate the compute units.
However, during the decoding phase, the workload often reduces to a thin GEMM with shape $(b\times h) \times (h\times ah)$, where $b \ll h$ and $a \ge 1$.
In such cases, the resulting CI drops to approximately $2b$ for FP8 and $b$ for BF16, which is much lower than the threshold needed to achieve peak throughput at realistic batch sizes.
Furthermore, matrix multiplication units are typically optimized for fixed-size tile shapes, and high hardware utilization is only attainable when input dimensions align with hardware-friendly boundaries, such as multiples of 128~\citep{lee2024debunkingcudamythgpubased}.

KV cache computations are another bottleneck.
Unlike linear layers, increasing the batch size does not improve CI in attention because each sequence in the batch maintains a separate KV cache.
For example, in a BF16 KV cache with grouped query attention (GQA) using $g$ groups, the CI is bounded by $g$ FLOPs per byte.
As a result, even in an ideal implementation, the maximum achievable throughput for querying the KV cache is limited by the product of memory bandwidth and CI.
In the case of an LLaMA v3.1 8B Instruct model with $g = 4$ running on a Gaudi 2, this translates to a theoretical upper bound of $2.4\,\text{TB/s} \times 4 = 9.6$~TFLOPS.
Because attention computations are inherently memory-bound and the number of FLOPs required for them scales linearly with sequence length, decoding becomes increasingly constrained by attention throughput as sequence lengths grow.
Recognizing the distinct compute characteristics of each phase enables more effective optimization strategies, particularly in improving decode phase efficiency, which often dominates end-to-end inference latency.

\section{Empirical Comparisons}\label{sec:comparisons}

\subsection{Prefill Phase}
\begin{figure}[ht]
    \centering
     \hspace{-0.5cm} % 왼쪽으로 1cm 이동
    \includegraphics[width=0.48\textwidth]{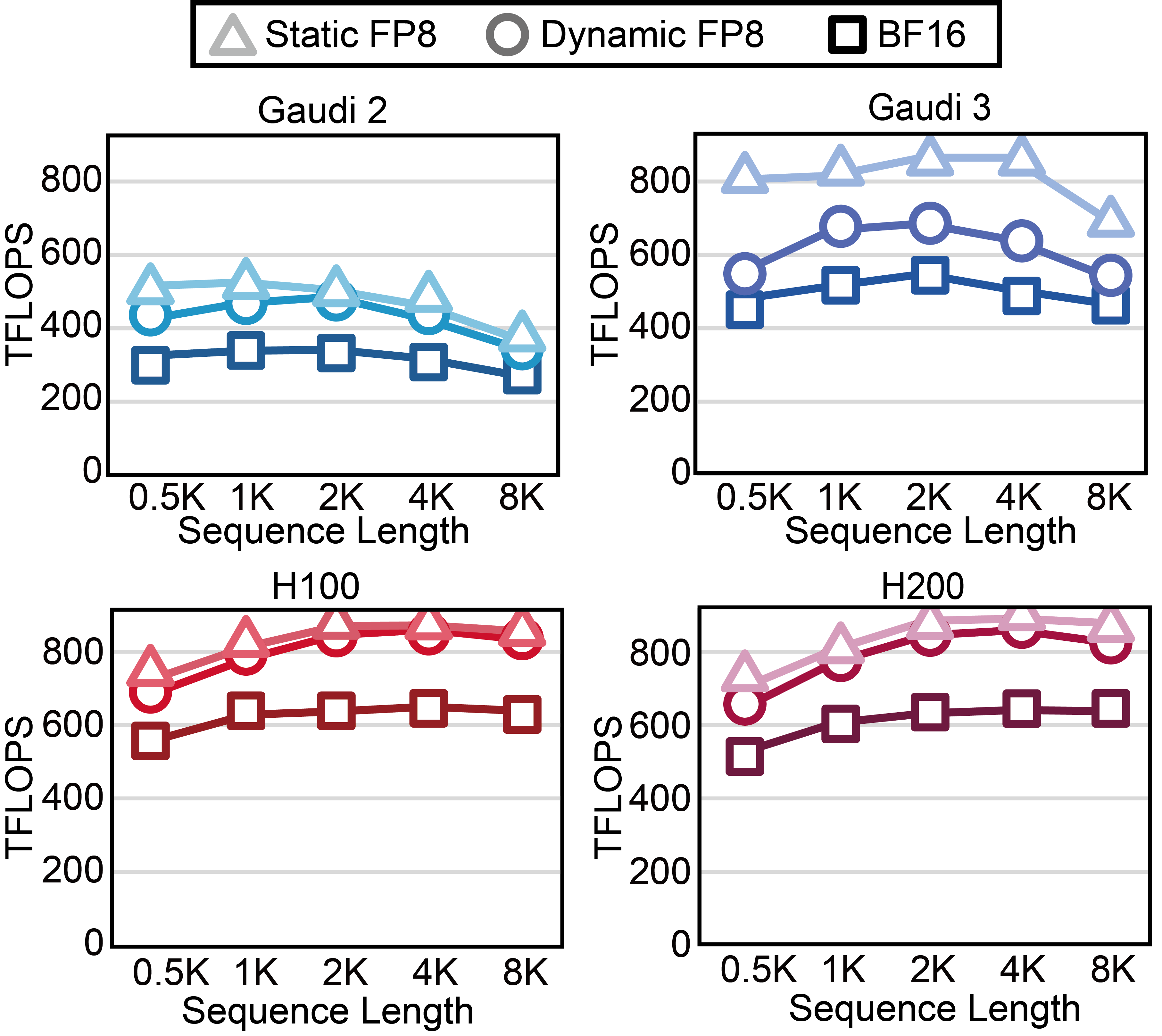}
    \caption{Diagram showing prefill phase throughput measurements for batch size 1. Throughput improves with longer sequence lengths until they begin to decline as the proportion of attention computations, which are slower than GEMMs, takes up a greater share of the computation.}
    \label{fig:prefill_roofline}
    % \vskip -0.1in
\end{figure}

\begin{figure}[ht]
     \centering
     \hspace{-0.5cm} % 왼쪽으로 1cm 이동
    \includegraphics[width=0.48\textwidth]{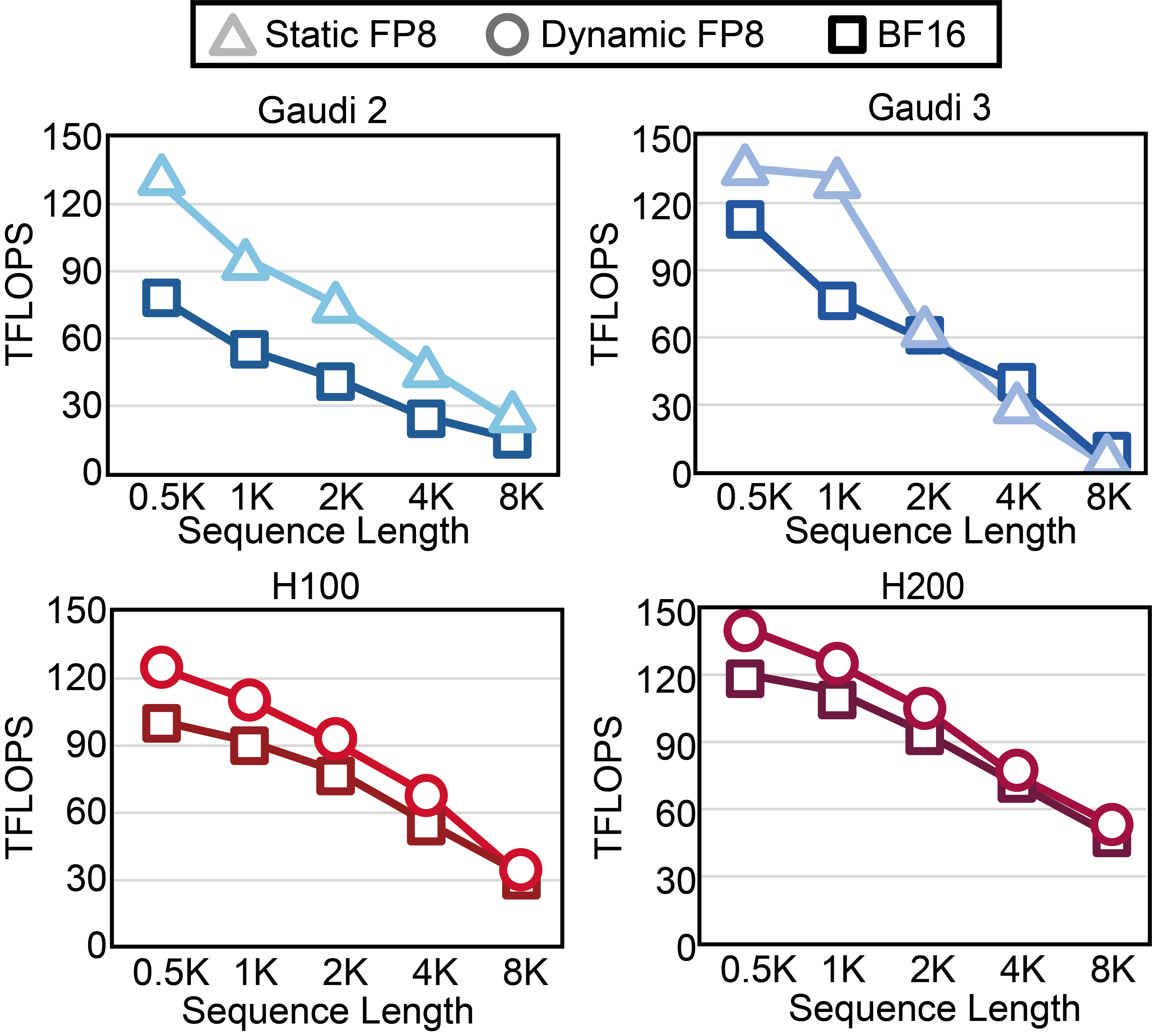}
    \caption{Decode throughput comparison between BF16 and FP8 on Llama v3.1 8B Instruct using a batch size of 64 using static FP8 scaling on the Intel HPUs (Gaudi 2 \& 3, top) and dynamic FP8 scaling on NVIDIA GPUs (H100 \& H200, bottom). Throughput differences between FP8 and BF16 in the Gaudi 2 are 50\% or greater, whereas they are under 25\% for the H100 and H200.}
    \label{fig:static_dynamic}
\end{figure}

Figure~\ref{fig:prefill_roofline} presents a comparison of prefill throughput, measured in TFLOPS, for the LLaMA v3.1 8B Instruct model across various input sequence lengths on the Gaudi 2, Gaudi 3, H100, and H200 for BF16, static FP8 scaling, and dynamic FP8 scaling. The results show that prefill performance is primarily determined by GEMM throughput rather than memory bandwidth. The H100 and H200 GPUs show a higher prefill throughput than the Gaudi 2, in line with their higher specifications. The Gaudi 3 has a lower throughput than the H200, despite having a peak BF16 GEMM throughput of 1678 TFLOPS \citep{gaudi3_whitepaper} compared to 989.4 TFLOPS for the H200 \citep{h200_datasheet}. We explore the reasons for this discrepancy in Section \ref{sec:insights}.

\begin{figure*}[ht]
    \centering
    \includegraphics[width=0.99\textwidth]{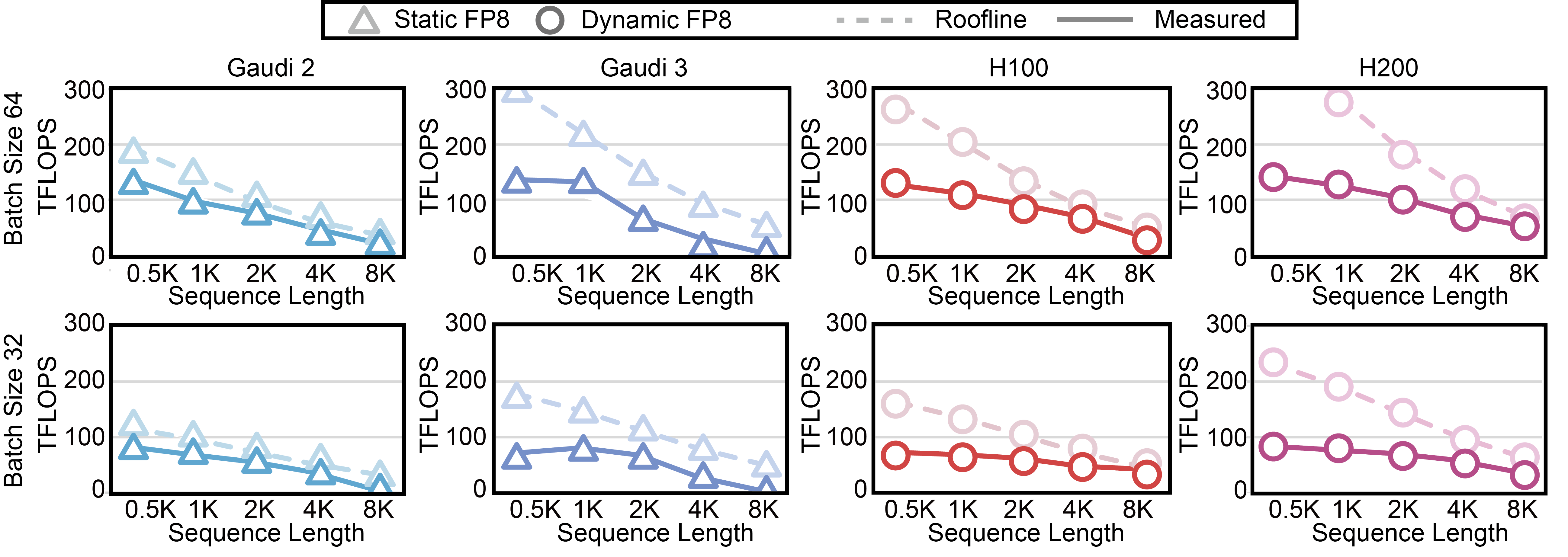}
    \caption{Comparison between measured and roofline throughputs obtained from hardware specifications across varying sequence lengths and batch sizes for the Llama v3.1 8B Instruct model.
    The dashed lines indicate the roofline estimates, whereas the solid lines represent the measured TFLOPS with FP8 quantization.
    }
    \label{fig:decode_roofline}
    \vskip -0.1in
\end{figure*}

All accelerators exhibit higher throughput when using FP8 compared to BF16.
For NVIDIA GPUs, static scaling yields slightly better performance than dynamic scaling, as dynamic scaling incurs an overhead of computing scaling factors during inference time.
The performance gap for Gaudi devices between static and dynamic scaling is larger, which we attribute to the immaturity of the dynamic scaling implementation, which was not feature-complete as of the time of writing.

\subsection{Decode Phase}

The decoding phase has become the primary bottleneck in end-to-end inference for autoregressive LLMs.
As recent "reasoning" models~\citep{deepseekai2025deepseekr1incentivizingreasoningcapability} often generate hundreds or even thousands of tokens to produce high-quality responses, the duration of the decode phase increasingly dominates total inference time.
Optimizing decoding is therefore essential to reduce latency and improve overall throughput.

Figure~\ref{fig:static_dynamic} presents a comparison of decode throughput across four accelerators. 
BF16 and static FP8 scaling are used on HPUs, whereas BF16 and dynamic FP8 scaling are compared for GPUs. 
While software optimization differences exist across devices, our comparison focuses on the best-performing configurations for each, specifically those showing higher TFLOPS than their respective BF16 baselines.
A key observation is that the Gaudi 2 achieves comparable or even higher throughput in FP8 than the BF16 throughput of the other devices, all of which have higher hardware specifications. 
This result underscores the effectiveness of adopting low-precision FP8 for memory-bound workloads, often yielding greater efficiency than relying solely on newer or higher-performance hardware.

However, the performance gains depend heavily on the underlying hardware architecture.
For example, Gaudi 2 shows a substantial FP8 improvement of over 65\%, while NVIDIA GPUs yield more modest gains of under 25\%. 
The Gaudi 3 also exhibits unstable FP8 throughput, performing well on short sequences but degrading sharply on longer ones.
These results suggest that fully leveraging FP8 requires co-design between hardware and software.

As shown in Figure~\ref{fig:decode_roofline}, the discrepancy becomes more evident when comparing theoretical roofline performance with actual throughput.
On the Gaudi 2, the measured throughput closely aligns with the theoretical roofline, whereas other accelerators exhibit noticeable gaps.
In practice, lower-end devices can match or even exceed the performance of higher-end counterparts using FP8, offering a significant TCO advantage.

%Surprisingly, the Gaudi 2 not only demonstrates superior power efficiency across all sequence lengths but also achieves higher absolute throughput for short sequences below 4K tokens. At longer sequence lengths, however, memory bandwidth becomes the primary bottleneck. Our measurements show HBM bandwidths of 2.0 TB/s for the Gaudi 2 and 2.6 TB/s for the H100, giving the latter an advantage as the KV cache size increases.

\section{Insights into Hardware Design}\label{sec:insights}
\subsection{LLM Decoding and Thin GEMM Throughput}

\begin{table*}[ht]
\small
\setlength{\tabcolsep}{14pt}
\centering
\caption{Thin GEMM throughputs (in TFLOPS) of the Gaudi 2, Gaudi 3, H100, and H200
}
\begin{tabular}{rrrrrrrrrr}
\toprule[1.2pt]
GEMM   TFLOPS            & \multicolumn{1}{c}{K, N} & \multicolumn{4}{c}{2048}   & \multicolumn{4}{c}{4096}    \\
\cmidrule{2-10}
Shape: (M,K,N)           & \multicolumn{1}{c}{M}    & 8    & 16   & 32   & 64    & 8    & 16   & 32    & 64    \\
\midrule
\multirow{2}{*}{Gaudi 2} & BF16 & 13.1 & 25.8 & 50.4 & 110.5 & 18.8 & 37.3 & 73.9  & 145.0   \\
                         & FP8  & 24.6 & 47.7 & 88.8 & 162.2 & 35.7 & 67.8 & 131.4 & 252.5 \\
\midrule
\multirow{2}{*}{Gaudi 3} & BF16 & 16.7 & 34.6 & 45.3 & 130.9 & 26.9 & 53.6 & 106.0 & 203.0 \\
                         & FP8  & 24.6 & 47.4 & 66.9 & 157.1 & 49.9 & 99.1 & 193.5 & 370.9 \\
\midrule
\multirow{2}{*}{H100}    & BF16 & 5.3  & 10.7 & 19.9 & 41.2  & 14.7 & 28.5 & 68.3  & 133.2 \\
                         & FP8  & 5.3  & 10.4 & 21.0 & 41.9  & 15.5 & 31.0 & 63.8  & 120.5 \\
\midrule
\multirow{2}{*}{H200}    & BF16 & 6.2  & 12.1 & 24.2 & 48.0  & 17.8 & 34.4 & 67.9  & 158.2 \\
                         & FP8  & 6.7  & 13.4 & 26.5 & 53.3  & 18.3 & 36.7 & 72.7  & 151.3 \\
\bottomrule[1.2pt]
\end{tabular}
\label{tab:thin_gemm}
\end{table*}

\begin{figure*}[ht]
     \hspace{-1cm} % 왼쪽으로 1cm 이동
    
    \includegraphics[width=1\textwidth]{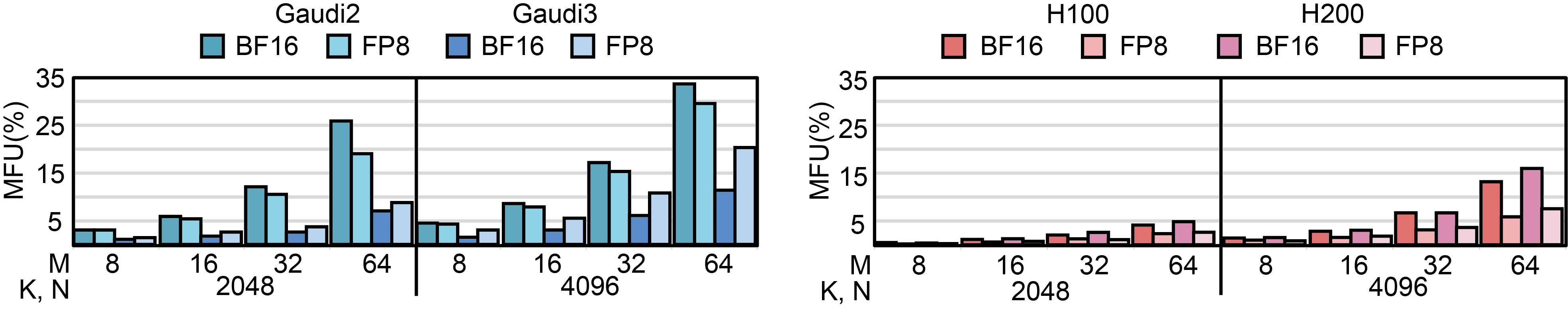}
     % \vspace{0.5em} % 그림과 캡션 사이 간격을 추가
    \caption{Thin GEMM MFU comparison for both BF16 and FP8. The Gaudi HPUs maintain a similar MFU for BF16 and FP8 of similar shapes, while there is a noticeable drop for the NVIDIA GPUs. The MFU differences at the same shapes are enough to provide superior TFLOPS for the Gaudi HPUs over the NVIDIA GPUs as shown in Table \ref{tab:thin_gemm}.}
    \label{fig:GEMV_MFU}
    \vskip -0.1in
\end{figure*}

% thin gemm이 llm decode phase의 중심이고 앞으로도 중요해질 것, 잘 분석해야 한다~
The dominant workload during LLM decoding is the thin GEMM, making thin GEMM throughput the dominant influence on overall inference latency. As shown in Equation \ref{eq:model_flops_delta_batch}, the decoding stage at short sequence lengths ($s < h$) is dominated by linear operations, while longer sequences are increasingly dominated by attention.
Although LLM sizes continue to scale, their hidden dimensions grow at a much slower rate. For example, Llama v3.1 8B has a hidden size of 4096, whereas Llama v3.3 70B increases it only to 8192. Tensor parallelism (TP), which is commonly employed to reduce latency and memory overhead, further partitions model weight matrices into smaller submatrices.
The growing adoption of fine-grained Mixture-of-Experts (MoE) models \citep{deepseekai2024deepseekv3technicalreport, llama4_blog} also contributes to this trend, as MoE models consist of a large number of experts with relatively small hidden dimension sizes.

% This superior efficiency of the Gaudi 2 for small matrices is particularly relevant given that LLM hidden dimensions typically range between 1K (\textit{e.g.}, Llama 1B) and 8K (\textit{e.g.}, Llama 70B), with tensor parallelism further reducing matrix sizes. Moreover, the lower power draw of the Gaudi 2 relative to its TDP, as shown in Table \ref{tab:gemm_tflops_power}, suggests that naïve TDP comparisons can be misleading, emphasizing the need for empirical evaluation.

% To investigate the lack of performance gains in the H100, we analyze its throughput on thin matrices, identifying it as the primary bottleneck. Table \ref{tab:thin_gemm} shows that while FP8 GEMM throughput on the Gaudi 2 is nearly twice that of BF16 GEMM, there is minimal improvement on the H100. 

Table \ref{tab:thin_gemm} presents GEMM TFLOPS in a setting where the dimension of the matrix $M$ is significantly smaller than $K$ and $N$. This setup is similar to the thin GEMM condition faced during decoding with a batch size of $M$. Row-wise scaling is used for the FP8 GEMMs. Quantization is excluded from the measurement. Both input matrices are in FP8 while the output matrix is in BF16.

Throughput scales linearly with $M$ on all devices, but Gaudi HPUs tend to achieve higher throughput than NVIDIA GPUs for both BF16 and especially FP8.
Compared to their performance on large GEMMs, NVIDIA GPUs show notably lower throughput than their specifications under these thin GEMM conditions. 
FP8 and BF16 throughputs for the NVIDIA GPUs are nearly identical, failing to reflect the expected performance gains from lower precision.
In contrast, Gaudi HPUs exhibit an FP8 throughput approximately double that of BF16, aligning well with expectations.

% thin gemm의 mfu 측정 방법 및 측정 결과

Figure \ref{fig:GEMV_MFU} presents the MFU of thin GEMM operations, computed using the method described above.
While the Gaudi\,3 exhibits some fluctuation, the overall MFU of Gaudi HPUs consistently exceeds that of NVIDIA GPUs. Especially in the case of the Gaudi 2, despite its having a lower peak TFLOPS and memory bandwidth than both the H100 and the H200, it has a higher absolute throughput for most of the matrix shapes in Table \ref{tab:thin_gemm}.
This indicates that HPUs achieve more effective performance in thin GEMM workloads primarily due to their higher MFU, which enables more efficient utilization of compute units for memory-bound computations. An interesting observation is that the Gaudi 3 has a lower MFU than the Gaudi 2, with the Gaudi 3 sometimes even having a lower absolute throughput as well. % We attribute this to the 

\subsection{MME Differences Between GPUs and HPUs}

Although FP8 offers up to double the throughput of BF16 or FP16, suboptimal input/output pipelining or memory access patterns can severely bottleneck overall throughput, even if the compute units are capable of fast execution. 
Unlike large matrix-matrix multiplications, thin GEMMs fail to fully utilize computing units. Moreover, the inefficiency that derives from hardware utilization increases as low data precision is adopted.
Therefore, to operate datacenters efficiently, it is crucial to understand how specific accelerators perform in scenarios involving thin GEMM and low-precision computations.

\begin{figure}[ht]
    \centering
    \includegraphics[width=0.5\textwidth]{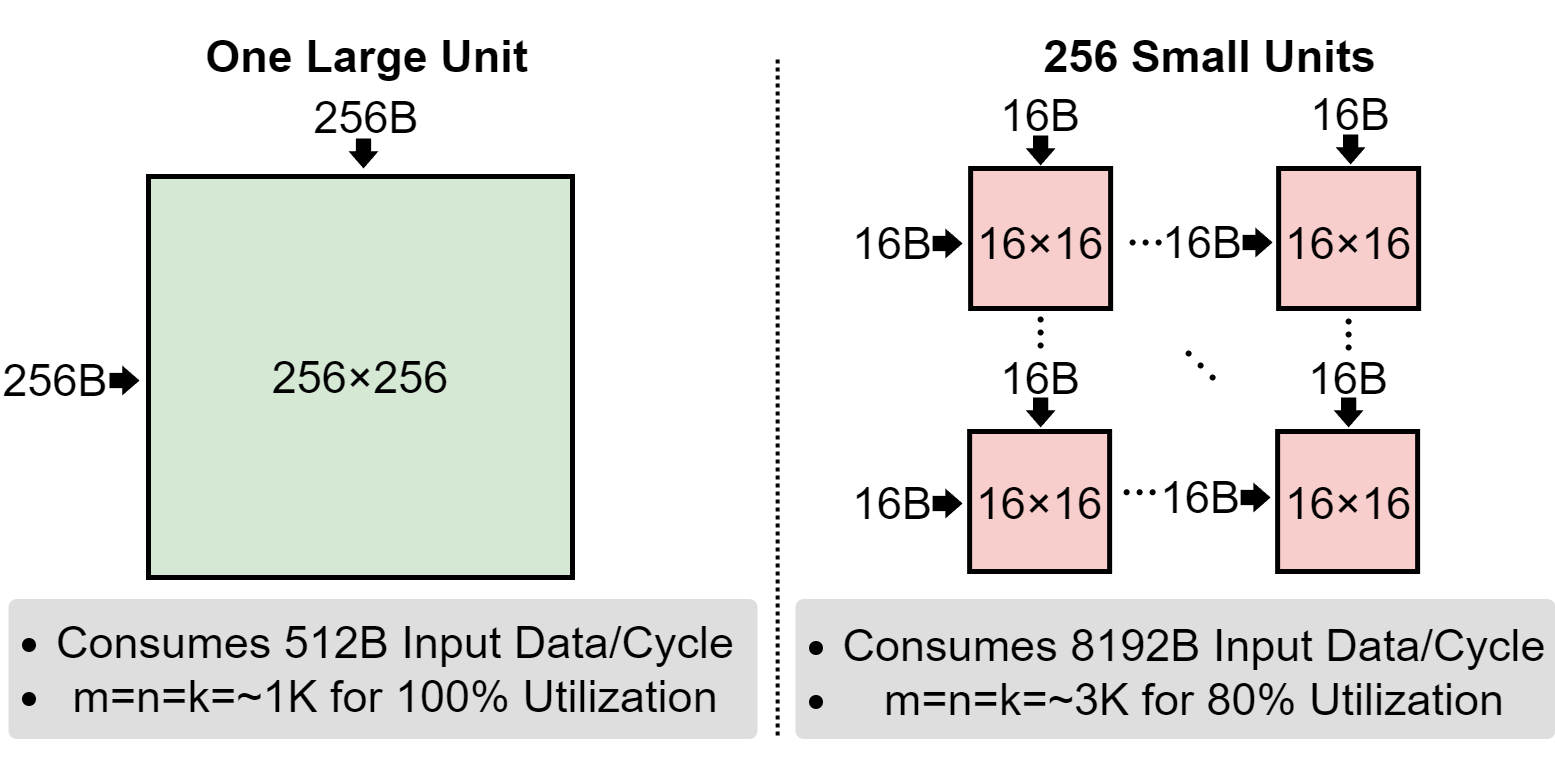}
    \caption{Gaudi HPUs employ a few large systolic array units in the MMEs, unlike NVIDIA GPUs, which employ many small units across streaming multiprocessors.}
    \label{fig:mme_architecture}
    % \vskip -0.1in
\end{figure}

The differences in FP8 thin GEMM performance are likely attributed to the architectural differences of the accelerators.
Figure \ref{fig:mme_architecture} depicts the differences between the MME architectures of Gaudi HPUs and those of other commercial hardware solutions. 
%In the case of NVIDIA Hopper, details of the Tensor Core architecture remain undisclosed, making it difficult to analyze the actual implementation. We expect that future disclosure of internal implementation details would enable a more systematic investigation. In contrast, Gaudi2 provides public documentation that contains their Matrix Multiplication Engine (MME) architecture, which is an output-stationary systolic array.
In architectures that rely on several small units, each unit consumes input data in every cycle, incurring significant overhead in unit-to-unit data movement.
In contrast, one large MME consumes a much smaller amount of input data per cycle and requires less expensive ALU-to-ALU communication within a whole MME.
 As a result, Gaudi HPUs require fewer input elements per cycle to fully utilize compute resources, thereby reducing first-level memory bandwidth requirements and improving efficiency \citep{gaudi2_whitepaper, gaudi3_whitepaper}.

\begin{figure}[ht]
    \centering
    \includegraphics[width=0.5\textwidth]{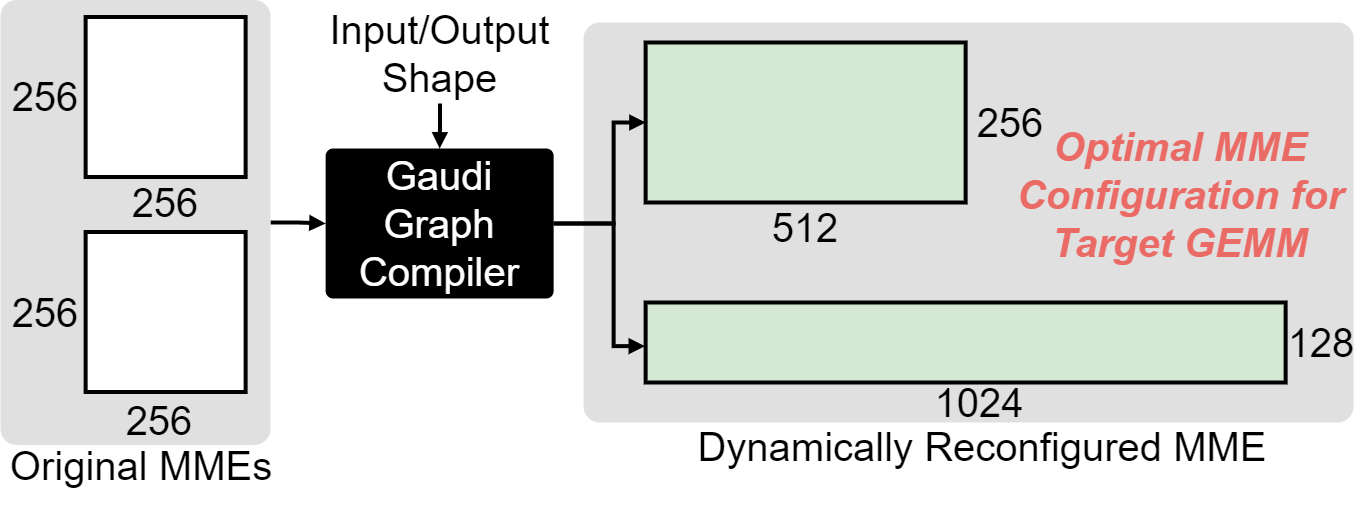}
    \caption{Dynamically reconfigurable MME architecture on Gaudi HPUs for target GEMM size optimization. The minimum size of the reconfigured MME unit is 128 elements for both BF16 and FP8.}
    \label{fig:reconfigurable_mme}
    % \vskip -0.1in
\end{figure}

Furthermore, Figure \ref{fig:reconfigurable_mme} illustrates the reconfigurable MME sizes of Gaudi HPUs.
Gaudi HPUs leverage a graph compiler to dynamically reconfigure the MME size based on the target GEMM dimensions, optimizing resource utilization \citep{lee2024debunkingcudamythgpubased}.
While the Gaudi HPU architecture contains MMEs with size $256 \times 256$, they can be folded and integrated in different sizes with a minimum width of 128. 
This flexibility allows the hardware to adapt to a wide range of GEMM dimensions in practice, offering substantial potential for improving LLM inference efficiency.

\subsection{Exponential Function Optimization}

While the Gaudi 2 \& 3 outperform the H100 and H200 in the LLM decode phase for short sequence lengths of below 512, their throughput gains diminish at longer sequence lengths. We find that this decline is primarily due to the absence of dedicated special function units (SFUs, also referred to as multi-function units, or MUFUs) in Gaudi HPUs to accelerate exponential calculations in the softmax.

Although GEMM operations that leverage tensor cores dominate LLM workloads, specialized functions such as softmax and SiLU are also necessary. During LLM inference, SFUs perform the exponential computations in these functions efficiently.
For decoding workloads, the ratio of exponential computations relative to GEMM computations increases in proportion to the sequence length. As a result, the softmax operation becomes a bottleneck for decoding at longer sequence lengths.
GPUs equipped with SFUs can process exponential functions on SFUs while simultaneously executing GEMMs on tensor cores to pipeline the workload.

In contrast, both the Gaudi 2 \& 3 have a disadvantage in that they lack SFUs and must handle exponential computations via their TPC cores. They also have a lower ratio of vector core compute relative to matrix computation compared to the H200. Peak TPC core BF16 throughput is 11 TFLOPS for the Gaudi 2 and 28.7 TFLOPS for the Gaudi 3, whereas the H200 has 133.8 TFLOPS BF16 CUDA core throughput and includes separate SFU cores.
Because of this, the softmax becomes a greater bottleneck for the Gaudi during decoding.
This implies that despite the notable efficiency of thin GEMMs in Gaudi HPUs, exponential function calculations bottleneck HPU decoding throughput for long sequences, as the softmax computation scales $O(BS)$ for batch size $B$ and sequence length $S$ during decoding.
% when an SFU is lacking, the specialized function must also be handled by the ALUs, resulting in reduced parallelism and a scheduling bottleneck.
% This issue is especially pronounced in attention blocks, which contain softmax operations. While H100 handles softmax via the SFU and keeps tensor cores occupied with GEMM, Gaudi2 incurs considerable overhead for softmax processing unless further optimizations are employed. 

Gaudi HPUs also incorporate an embedded look-up table (LUT), which could store precomputed special function values via piecewise linear approximation. If the LUT offers sufficiently low access latency and can hold a large enough range of values, it may be a valid substitute for SFUs.
% However, Gaudi 2 currently has no software stack to program the LUT. Although Gaudi 2 has strong potential to improve performance in long sequence length, the software stack is limiting the practical effectiveness.
However, Gaudi HPUs currently lack software support for easily programming the LUTs.
Although it has potential in improving performance with long sequences, the limited software stack constrains its practical effectiveness.
% Although Gaudi 2 has strong potential to improve performance in long sequence length, the current software lacks the necessary programmability, limiting the practical effectiveness.
% Despite its potential to mitigate the softmax bottleneck, the current software stack lacks the necessary programmability to fully utilize this hardware feature.

% PCIe power limit CPU/GPU => Called power steering by NVIDIA.

% \subsection{Chiplet Interposer Power Consumption}

% The Gaudi 3 uses a chiplet \citep{chiplet} architecture where two chips are connected via an interposer. This functions to increase the HBM memory capacity and bandwidth by allowing more HBM stacks to be read simultaneously. However, we find that 

\subsection{Power Capping}

Our experiments in Section \ref{sec:comparisons} showed that although the Gaudi 2 could often outperform the H100 and H200 on memory-bound thin GEMMs, the Gaudi 3 sometimes even fell behind the Gaudi 2. An investigation into the root cause found that the Gaudi 3 had low throughput because of power throttling within the chip to adhere to the TDP constraints. On closer inspection, we discovered that the Gaudi 3 reached its TDP of 900W even for thin GEMMs. The reason that it had a lower prefill throughput than the H200, despite having a higher peak TFLOPS, was also because power throttling reduced the achieved GEMM throughput of the Gaudi 3.

Although setting the power cap too low as in the case of the Gaudi 3 can cause problems due to having too little power available, prior work \cite{zhao2023sustainable} has found that power capping at appropriate levels can improve hardware reliability and mean time between failures. Power capping can also reduce infrastructure costs by allowing more servers to be placed in a single datacenter rack. Therefore, power capping is a viable strategy if the throughput decrease is less than the TCO gain.

\begin{table}[ht]
\centering
\small
\caption{Decode TFLOPS with and without a 400W cap on the H200}
\begin{tabular}{c|ccc|ccc}
\toprule[1.2pt]
        & \multicolumn{3}{c|}{\textbf{Llama v3.1 8B (BS=64)}} & \multicolumn{3}{c}{\textbf{Llama v3.3 70B (BS=4)}} \\
\textbf{Length} & \textbf{W/O}     & \textbf{W/}      & \textbf{Difference}    & \textbf{W/O}      & \textbf{W/}     & \textbf{Difference}     \\
\midrule
512     & 141     & 132     & 7\%             & 23       & 19     & 19\%             \\
1024    & 126     & 111     & 11\%            & 23       & 19     & 17\%             \\
2048    & 105     & 86      & 18\%            & 24       & 19     & 21\%             \\
4096    & 76      & 64      & 15\%            & 21       & 19     & 9 \%             \\
8192    & 56      & 40      & 28\%            & 23       & 18     & 21\%            \\
\bottomrule[1.2pt]
\end{tabular}
\label{tab:H200_powercap}
\end{table}

Table~\ref{tab:H200_powercap} compares the decode throughput of the H200 across varying sequence lengths under its default 700W TDP and when applying a power cap of 400W for Llama 8B and 70B models with batch sizes of 64 and 4, respectively. Although previous works such as SplitWise \citep{splitwise} found that power capping did not slow decoding at 400W, we find that the same power cap causes noticeable throughput degradation, possibly because framework updates have made LLM decoding more efficient.

A limitation of currently available power capping solutions is that they operate on a single card and cannot share their cap. However, considering that the primary purpose of power capping is to limit the peak power consumption of the rack, it would be beneficial to be able to pool the cap across many devices. However, this functionality is only available in Grace superchips with automatic power steering capabilities.

\section{Throughput Shifts the TCO Balance}

\subsection{Hardware Characteristics Related to Data Precision} 
We now summarize how hardware support for low-precision computation influences the cost-efficiency of LLM inference, as captured by our TCO model.
In the decoding phase of LLM inference, which often dominates total latency, Gaudi 2 achieves nearly 50\% higher throughput using FP8 quantization than BF16, while the H100 and H200 show little to no improvement. This results in an upward shift along the y-axis in the TCO table shown in Figure~\ref{fig:TCO_variation}, moving the evaluation point into the green region, which indicates the improved cost efficiency of the Gaudi 2. Conversely, in long-sequence scenarios, the Gaudi 2 \& 3 suffer from softmax bottlenecks due to the absence of a dedicated SFU or alternative workaround, reducing $R_{Th}$ and shifting the evaluation point downward into the red region, where choosing H100 and H200 becomes more cost-effective solution. 

\subsection{Workload Characteristics} 
In practical deployment, inference services must satisfy diverse service-level objectives (SLOs) depending on the application. For example, latency-sensitive workloads such as real-time voice dialogue systems require near-instantaneous responses, making low-latency SLOs essential. In contrast, creative generation tasks such as image or slide synthesis can tolerate longer delays in exchange for higher output quality. These differences in performance requirements directly affect achievable throughput, thus resulting in different vertical positions on the TCO table.

To support such a variety of workloads, datacenters are required to deploy inference systems that can efficiently serve both lightweight and large-scale models. SLO-aware inference system optimization frameworks, such as \cite{huang2025slo, cheng2025scoot}, dynamically adjust resource allocation and engine parameters to handle multi-SLO environments. As a result, even for the same hardware configuration, the evaluation point on the TCO table may shift depending on the workload.

In this context, hardware selection is guided not by a single evaluation point, but by the overall distribution of expected workloads. Since each application maps to a unique point on the TCO surface, robust deployment decisions require analyzing these points collectively. The optimal hardware is the one that maintains strong cost-efficiency across the majority of operating scenarios, rather than excelling in only a narrow performance region.

\section{Limitations}

In this study, we present a scalable framework for comparing AI accelerator options in terms of datacenter TCO. However, our study has several limitations.

First, we did not discuss KV cache quantization \citep{oaken, yang2024tokenleftbehindreliable}, as this is a memory-saving technique, distinct from GEMM computation. While KV cache quantization can improve decode throughput by reducing the memory read from DRAM, the GEMM computations within the attention operation are conducted in BF16 to preserve accuracy, even in models such as DeepSeek-V3 \citep{deepseekai2024deepseekv3technicalreport} where linear layers use FP8 quantization.

Second, we concentrate on evaluating performance on a single device without factoring in the communication costs between devices or nodes. This simplifies the analysis because disaggregated inference allows devices to focus on a single workload. 
However, it limits our analysis of tensor parallelism, which is dependent on low-latency communications. For example, tensor parallelism can be made more efficient by asynchronously overlapping the GEMM computations with the buffer communications \citep{asynctp}, but asynchronous tensor parallelism is unavailable for the Gaudi as of the time of writing.

Third, our analysis focuses on Llama-based dense models \citep{grattafiori2024llama3herdmodels}. As such, the results may not directly generalize to other workloads, particularly those involving MoE inference. We intentionally simplify our analysis to focus on a single device with a uniform workload, even though workload imbalances at both the node and device levels are a critical issue in MoE inference \citep{deepseekv3_insights}, which requires sophisticated load balancing for the experts. However, analyzing complex workloads requires breaking them down into their components, which our work has attempted to do. Moreover, as efficient MoE inference requires disaggregation \citep{deepseekai2024deepseekv3technicalreport}, our analysis of the component workloads serves as a starting point for further investigation into more sophisticated deployments. Also, note that MoE inference reduces the average batch size of each expert by the ratio of active experts to the total number of experts, heightening the importance of thin GEMM analysis.

\section{Conclusion}

% In this study, we present a scalable framework for comparing AI accelerator options in terms of datacenter TCO.
% Using our model, we examine two key factors that influence throughput and are directly connected to the TCO: hardware characteristics related to the numerical precision and workload characteristics % specific to the LLM inference. We evaluate the representative next-generation accelerators, the NVIDIA H100, H200, and Intel Gaudi 2, Gaudi 3, and analyze their behavior under FP8 quantization.
% 
% H
% From the perspective of CSPs who deploy LLM services, energy efficiency is a more practical and impactful metric. 
% In this context, a software-aware approach, especially leveraging low-precision data types such as FP8, offers a more cost-effective and scalable strategy for next-generation LLMs.

The incredible progress in LLM capabilities seen over the last few years has ushered in an unprecedented buildup of datacenter capacity as corporations and nation states spend hundreds of billions of dollars to hasten the adoption of AI. 
In this study, we present a scalable framework for comparing AI accelerator options in terms of datacenter TCO. 
Using our model, we examine two key factors that influence throughput and are directly connected to the TCO: hardware characteristics related to numerical precision and workload-specific behavior in LLM inference. We evaluate four AI accelerators: the Intel Gaudi 2 \& 3 and the NVIDIA H100 \& H200, and analyze their behavior under FP8 quantization.

Recent AI accelerators primarily focus on maximizing raw computational performance, typically measured in peak TFLOPS. However, our study highlights that such a singular emphasis may result in inefficient AI datacenter management. From the perspective of CSPs who deploy LLM services, throughput measurements on their target workloads, as well as considerations such as power efficiency, are also critical.
We highlight the key role of thin GEMMs and FP8 quantization in LLM decoding, a highly important aspect of optimizing AI datacenter TCO. 
% We hope that the insights provided in this work help inform deployment decisions and guide future accelerator design efforts aimed at improving LLM inference in datacenter environments.
We hope that the insights provided in this work will inform practical deployment decisions and guide future accelerator design efforts toward more efficient and workload-aware LLM inference at scale.
% In this study, we present a scalable framework for comparing AI accelerators in terms of datacenter TCO, focusing on two key dimensions: hardware characteristics related to numerical precision and workload-specific behavior in LLM inference.
% To this end, we propose FP8 as a practical baseline data precision for next-generation LLMs and analyze its implications across representative accelerators: NVIDIA H100, H200, and Intel Gaudi 2, Gaudi 3.
% Our analysis extends beyond FP8-based inference at the decode phase, where thin GEMM dominates. 
% We demonstrate that hardware support for FP8, coupled with LLM workload-aware design, is highly important for optimizing datacenter TCO.

% While our study provides valuable insights into cost-performance trade-offs, additional research is required to generalize the results to full-scale datacenter environments with heterogeneous infrastructure, workload diversity, and varying deployment strategies.

% \section*{Acknowledgements}
% This document is an updated version of HPCA 2022 and 2023, which, in
% turn, has been derived from two previous conferences, in particular
% HPCA 2021 and MICRO 2021, which, in turn, are derived from past MICRO,
% HPCA, ISCA, and ASPLOS conferences.

%%%%%%% -- PAPER CONTENT ENDS -- %%%%%%%%

%%%%%%%%% -- BIB STYLE AND FILE -- %%%%%%%%
% \bibliographystyle{IEEEtranS}
\bibliographystyle{unsrtnat}
\bibliography{refs}
%%%%%%%%%%%%%%%%%%%%%%%%%%%%%%%%%%%%

% \input{10_revision_letter}
\onecolumn
\newpage
\appendix
\section{Appendix}\label{sec:appendix}

\subsection{Differences in hardware capabilities}

We compare implementations of FP8 GEMM between NVIDIA GPUs and Intel Gaudi HPUs, identifying key differences despite both adhering to the FP8 specification \citep{micikevicius2022fp8formatsdeeplearning}.

\textbf{(Accumulation precision)} Hopper GPUs use a 14-bit accumulator for FP8 GEMMs \citep{deepseekai2024deepseekv3technicalreport}, requiring casting to CUDA cores for higher precision. Software optimizations, such as applying FP32 accumulation to only one in four WGMMA instructions, reduce error but increase kernel complexity and remain less precise than full FP32 accumulation. In contrast, Gaudi HPUs always accumulate in FP32 \citep{lee2025fasterinference}, ensuring higher numerical precision.

\textbf{(E4M3 range)} The Gaudi 2 follows the IEEE specification for special values, unlike NVIDIA GPUs, which use a single special value representation \citep{noune20228bitnumericalformatsdeep}. This results in seven fewer magnitude representations and a maximum value of 240 for E4M3 in the Gaudi 2 compared to 448 on NVIDIA GPUs. This has been addressed in the Gaudi 3, but we were unable to test these in our experiments.

\textbf{(Power-of-2 scaling)} Gaudi HPUs allow the modification of floating-point exponent biases to accelerate scaling factor application. The Gaudi 2 supports fixed hardware-accelerated scaling factors of  $2^{-8}, 2^{-4}, 2^{0}, 2^{4}$ for E4M3, while the Gaudi 3 extends this to arbitrary powers of 2 between $2^{-32}$ and $2^{31}$. However, this feature is limited to GEMMs with per-tensor scaling factors.

\textbf{(Stochastic rounding)} During FP8 quantization, stochastic rounding can be applied when converting 16/32-bit floating-point values to FP8, similar to the technique proposed for FP32-to-BF16 conversion in \citep{hlat}. This method is distinct from stochastic rounding applied at the inner-product \citep{doi:10.1137/22M1510819}.

\textbf{(Sparsity)} NVIDIA GPUs support semi-structured sparsity acceleration, potentially doubling tensor core peak throughput. However, despite attempts to leverage sparsity in LLMs \citep{sparsegpt, wanda_sun2024a}, dense GEMMs remain dominant due to accuracy loss and limited speedups. Gaudi HPUs do not support sparsity acceleration.

\begin{table}[h]
\small
\centering
\caption{Throughput and power measurements for square FP8 GEMMs with row-wise scaling. Ratios of measured TFLOPS and power to their peak values are shown in parentheses. H100 has peak FP8 throughput of 1989.9 TFLOPS with a TDP of 700W, and Gaudi 2 has 865 TFLOPS with a TDP of 600W.}
% \vskip 0.15in
\begin{tabular}{@{}ccrcc@{}}
\toprule
Device & (M,K,N) & \multicolumn{1}{c}{TFLOPS} & \multicolumn{1}{r}{Power (W)} & TFLOPS/W \\ \midrule
\multirow{4}{*}{Gaudi 2} & 1K & 367.9  (42.5\%) & 375 (63\%) & 1.0 \\
                         & 2K & 586.2  (67.8\%) & 460 (77\%) & 1.3 \\
                         & 4K & 817.1  (94.5\%) & 460 (77\%) & 1.8 \\
                         & 8K & 741.8  (85.8\%) & 490 (82\%) & 1.5 \\ \midrule
\multirow{4}{*}{H100}    & 1K & 218.3  (11.0\%) & 350 (50\%) & 0.6 \\
                         & 2K & 879.7  (44.2\%) & 690 (99\%) & 1.3 \\
                         & 4K & 1167.6 (58.7\%) & 690 (99\%) & 1.7 \\
                         & 8K & 1084.7 (54.5\%) & 690 (99\%) & 1.6 \\ \bottomrule
\end{tabular}
% \vskip -0.1in
\label{tab:gemm_tflops_power}
\end{table}

\begin{table}[h]
\small
\caption{Gaudi 2 throughput in TFLOPS for scaled FP8 GEMM for square matrices of the given size, excluding quantization overhead. Measured throughput relative to the peak FP8 throughput (865 TFLOPS) is included in parentheses. Hardware acceleration is only available for per-tensor scaling.}
\centering
\begin{tabular}{@{}ccllllll@{}}
\toprule
Type &
Size &
\multicolumn{1}{c}{Per-Row} &
\multicolumn{1}{c}{Per-Tensor} &
\multicolumn{1}{c}{HW Accel.} \\ \midrule
\multirow{4}{*}{E4M3} & 1K & 494 (57.1\%) & 494 (57.1\%) & 494 (57.1\%) \\
                      & 2K & 506 (58.5\%) & 641 (74.1\%) & 641 (74.2\%) \\
                      & 4K & 735 (84.9\%) & 796 (92.1\%) & 801 (92.6\%) \\
                      & 8K & 742 (85.7\%) & 822 (95.0\%) & 852 (98.4\%) \\ \midrule
\multirow{4}{*}{E5M2} & 1K & 306 (35.4\%) & 494 (57.1\%) & 493 (57.0\%) \\
                      & 2K & 506 (58.5\%) & 642 (74.2\%) & 642 (74.2\%) \\
                      & 4K & 735 (84.9\%) & 802 (92.7\%) & 802 (92.7\%) \\
                      & 8K & 726 (83.9\%) & 825 (95.4\%) & 825 (95.4\%) \\ \bottomrule
\end{tabular}
\label{tab:gaudi2_fp8_tflops}
\end{table}

\begin{table}[ht]
\small
\caption{H100 throughput (TFLOPS) for scaled FP8 GEMMs on square matrices of the given size, excluding quantization overhead. Ratios relative to peak FP8 throughput are shown in parentheses. Only E4M3 results are reported, as Python functions for E5M2 GEMM are unavailable for GPUs.}
\centering
\begin{tabular}{@{}ccrrrr@{}}
\toprule
Accum. & Size & \multicolumn{1}{c}{Per-Row} & \multicolumn{1}{c}{Per-Tensor} \\ \midrule
\multirow{4}{*}{FP32} & 1K & 217  (11.0\%) & 186 \:  (9.4\%) \\
                      & 2K & 299  (15.1\%) & 840  (42.4\%) \\
                      & 4K & 362  (18.3\%) & 1099 (55.5\%) \\
                      & 8K & 396  (20.0\%) & 1300 (65.7\%) \\ \midrule
\multirow{4}{*}{Fast} & 1K & 237  (12.0\%) & 147 \: (7.4\%) \\
                      & 2K & 810  (40.9\%) & 896  (45.3\%) \\
                      & 4K & 1136 (57.4\%) & 1205 (60.9\%) \\
                      & 8K & 1123 (56.8\%) & 1388 (70.1\%) \\ \bottomrule
\end{tabular}
\label{tab:h100_fp8_tflops}
\end{table}

\subsection{Stochastic rounding}

Hardware-accelerated stochastic rounding for FP8 quantization is a unique feature of Gaudi HPUs, inspired by techniques such as those proposed in~\citep{hlat}, and is not supported on NVIDIA GPUs.
Equation  \ref{eq:stochastic_rounding} formalizes the concept, where a higher-precision value  $x$ is rounded up to $x_{up}$ or down to $x_{down}$ stochastically based on the distance from $x$. 
\begin{equation}
  x_{quantized} =
    \begin{cases}
      x_{up}   \quad (p=\frac{x-x_{down}}{x_{up}-x_{down}}) \\
      x_{down} \quad (p=\frac{x_{up}-x}{x_{up}-x_{down}}).
    \end{cases}
\label{eq:stochastic_rounding}
\end{equation}

Stochastic rounding is expected to preserve more of the original numerical information post-quantization, potentially leading to improved model accuracy.
However, empirical results in Table \ref{tab:e5m2_sr} indicate that stochastic rounding during quantization does not significantly enhance model accuracy. Furthermore, it may even lead to accuracy degradation in certain cases \citep{hlat}. These findings suggest that while stochastic rounding during quantization theoretically retains more information, its practical benefits for FP8 inference in LLMs remain inconclusive.

\begin{table}[ht]
\centering
\caption{Comparison between different FP8 data types and rounding modes for MMLU CoT 5-shot performance on instruction-tuned Llama models. Stochastic rounding provides little or no benefit to accuracy while E5M2 is detrimental, especially for smaller models. Experiments were conducted on a Gaudi 2 HPU.}
% \vskip 0.15in
\small
\begin{tabular}{@{}lccc@{}}
\toprule
\multicolumn{1}{l}{\textbf{Model}} & \textbf{Data Type} &  \textbf{Rounding} & \textbf{MMLU} \\ \midrule
\multirow{3}{*}{Llama v3.2}  & BF16 & -   & 46.3\% \\
\multirow{3}{*}{1B Instruct}              & E4M3 & SR  & 45.7\% \\
                                        & E4M3 & RTN & 45.5\% \\
                                        & E5M2 & RTN & 44.5\% \\ \midrule
\multirow{3}{*}{Llama v3.2}  & BF16 & -   & 61.8\% \\
\multirow{3}{*}{3B Instruct}          & E4M3 & SR  & 61.7\% \\
                                        & E4M3 & RTN & 61.6\% \\
                                        & E5M2 & RTN & 60.7\% \\ \midrule
\multirow{3}{*}{Llama v3.1}  & BF16 & -   & 68.8\% \\
\multirow{3}{*}{8B Instruct}           & E4M3 & SR  & 68.3\% \\
                                        & E4M3 & RTN & 68.3\% \\
                                        & E5M2 & RTN & 67.5\% \\ \midrule
\multirow{3}{*}{Llama v3.3} & BF16 & -   & 82.0\% \\
\multirow{3}{*}{70B Instruct}              & E4M3 & SR  & 82.0\% \\
                                        & E4M3 & RTN & 82.0\% \\
                                        & E5M2 & RTN & 82.2\% \\ \bottomrule
\end{tabular}
% \vskip -0.1in
\label{tab:e5m2_sr}
\end{table}

\subsection{E4M3 vs. E5M2}

Prior work has shown that E4M3 yields better accuracy than E5M2 on language tasks~\citep{MLSYS2024_dea9b4b6}. We extend this analysis to instruction-tuned models, as summarized in Table~\ref{tab:e5m2_sr}. To identify the optimal format for inference, we compare the two using MMLU accuracy measured with LM Evaluation Harness (v0.4.7~\citep{eval-harness}). Across all evaluated models, E4M3 consistently outperforms E5M2. Moreover, Table~\ref{tab:gaudi2_fp8_tflops} shows that GEMM throughputs for both formats are comparable, making E4M3 the preferred choice despite its smaller representational range on the Gaudi 2.

\end{document}